# Unsupervised Classification of Intrusive Igneous Rock Thin Section Images using Edge Detection and Colour Analysis

S. Joseph

*Jabatan Mineral dan Geosains (Sarawak),*
*Kementerian Sumber Asli dan Alam Sekitar,*
*Kuching, Sarawak, Malaysia*
*silvia@jmg.gov.my*

H. Ujir and I. Hipiny

*Faculty of Computer Science & Information Technology*
*Universiti Malaysia Sarawak*
*Kota Samarahan, Sarawak, Malaysia*
*{uhamimah & mhihipni}@unimas.my*

***Abstract*—Classification of rocks is one of the fundamental tasks in a geological study. The process requires a human expert to examine sampled thin section images under a microscope. In this study, we propose a method that uses microscope automation, digital image acquisition, edge detection and colour analysis (histogram). We collected 60 digital images from 20 standard thin sections using a digital camera mounted on a conventional microscope. Each image is partitioned into a finite number of cells that form a grid structure. Edge and colour profile of pixels inside each cell determine its classification. The individual cells then determine the thin section image classification via a majority voting scheme. Our method yielded successful results as high as 90% to 100% precision.**

***Keywords— Minerals; Classification; Igneous Rocks; Edge Detection; Colour Analysis.***

## I. INTRODUCTION

Petrographic thin section microscopy study has a long tradition both in applied and academic geosciences. A geoscientist can straightforwardly identify a rock's constituent mineral phases, quantify fabric parameters and infer a rock's genesis using thin sections and a petrographic microscope.

Point counting of petrographic thin sections is the standard method for mineral and rock classification. This process is performed by a petrologist to find out the percentage of each mineral inside the thin image. Since this process is done manually, it is more inclined towards qualitative analysis rather than quantitative, as individual observation may vary and this could possibly lead to misclassification by the observer [1].

In mineralogy, a microscope is commonly used tool for manual mineral classification in thin sections. Experts have problems with automation of mineral classification process using a microscope [2]. Images obtained from rock thin sections exhibit inhomogeneous colours for each mineral type. This is caused by subtle tone changes which are related to inhomogeneous chemical composition and deformation due to stress affecting the rock sample [3].

There are inherent drawbacks of using petrographic microscopy. The process itself is time-consuming and iterative. It requires a human expert with substantial knowledge and experience in combining multiple petrographic and microscopy classification criteria, e.g., texture, colour, cleavage, twinning and tartan pattern, to perform the point counting. Due to this, we argue that an unsupervised classification of the rock thin sections to replace the human operator is therefore necessary.

This paper presents our unsupervised classification method and the classification results using edge and colour as the discriminating features.

## II. RELATED WORKS

According to [4], minerals comprising of igneous rocks are grouped either as primary or secondary minerals. Primary minerals are the type of minerals that crystallized directly from magma. According to [5], the composition of magma ranges widely. These compositional variations, together with rock textures, provide the best basis on which to classify igneous rocks and to distinguish them from other rock types

TABLE I.
MINERAL COMPOSITION OF INTRUSIVE IGNEOUS ROCKS, BASED ON THE QAPF DIAGRAM [6].

| No | Intrusive igneous rocks | Quartz % | Alkali Feldspar % | Plagioclase Feldspar % | Accessory Mineral % |
|---|---|---|---|---|---|
| 1 | Granite | 20-60 | 35-90 | 10-65 | 5-20 |
| 2 | Adamelite / Quartz monzonite | 5-20 | 35-65 | 35-65 | 10-35 |
| 3 | Tonalite | 15-50 | 10-35 | 65-90 | 10-40 |
| 4 | Diorite | <5 | - | 70-90 | 20-50 |

The 3 principal discriminant features used to classify igneous rocks are modal parameters, grain size characteristics and chemical characteristics. In this study, classification is solely based upon the determination of modal parameters. The Quartz, Alkali feldspar, Plagioclase, Feldspathoid (Foid) (QAPF) diagram is used in classification and nomenclature of coarse-grained crystalline rocks [6]. Percentage distribution of the modal mineral (QAPF) contents is shown in Table 1.

The optical properties of rock-forming minerals serve as the discriminative criteria for the classification of igneous rock thin section images. External knowledge on minerals with crossed polarizing filters, texture and colour is shown in Table 2.

TABLE 2.
DIAGNOSTIC MICROSCOPIC CHARACTERISTICS OF QUARTZ, ACCESSORY MINERAL, PLAGIOCLASE FELDSPAR & ALKALI FELDSPAR UNDER CROSSED POLARIZING FILTERS LIGHT [6].

| Bil | Mineral | Texture (medium/coarse grained) | Colour |
| --- | --- | --- | --- |
| 1 | Quartz | Generally unaltered and lacks visible twinning or cleavage. Clear surface. | Clear, milky white/ yellowish/ colourless. |
| 2 | Accessory Mineral | Most of the Accessory mineral are lacks twinning | Most of the Accessory mineral are colourful |
| 3 | Alkali Feldspar | Cross-hatched twinning or 'tartan' pattern | Medium gray |
| 4 | Plagioclase Feldspar | Repeat twinning /Lamellar twinning is visible in most of the crystals. Tabular crystal | Greyish |

Additionally, textural and compositional analyses of rock images can also be assessed using MastCam and RMI cameras [14]. The use of these specialised cameras allows easy distinction of the grain shape distribution.

### A. *Rock petrographic thin section image analysis*

In [7], a numerical approach is proposed based on image processing and multi-layer perceptron neural network that classifies carbonate rocks by grey level images digitized from thin section using Dunham textural classification. In their study, type of texture is predicted out from the digitized images of thin sections. [8] in their study of non-homogenous texture, they extract several features that characterize the colour and texture content. In their experiment, the feature vector of each sample is formed by the separate classification results. The final classification is made using the feature vector. As their findings, the non-homogenous textures can be classified effectively. Rocks petrographic thin section colour analysis

Colour pixels is used as the input for Artificial Neural Network (ANN) in mineral classification [2]. They compared RGB, L*a*b* and HSV colour spaces in colour texture analysis and used ten minerals for classification using colour and texture parameters using ANN. The classification results show that images of L*a*b* are not as accurate as the result of RGB images. [9] employed the statistical texture analysis methods for coloured textures. Multidimensional co-occurrence matrix for coloured texture images is calculated.

Gabor filters are applied to colour channels of rock's images in [8]. Using this method, the rocks can be classified using multiple scales and orientations. Two sets of industrial rock plate images have been used as testing material in the experiments. From the result, it shows that by using the colour information, classification accuracy is improved compared to conventional grey level texture filtering. The idea behind this approach is that the colour information of rock's images with texture description was suggested and developed. A typical characteristic of colour distribution is used in the image classification, and it can be described using statistical methods. [10] used colour moments to describe image colour distribution.

A classification of rock images using textural analysis is proposed in [15]. The textural feature is extracted from rock images. Some of the non-homogenous textural features are used to distinguish between the rock textures. For classification, K-NN classifier is used.

This paper focuses on the classification of two mineral types, i.e., quartz and accessory minerals. We decided to exclude plagioclase feldspar and alkali feldspar due to their non-homogenous texture. Furthermore, both plagioclase feldspar and alkali feldspar requires the thin section image to be captured in multiple stage rotation's angles. Different extinction angles (when viewed through a cross polarizer microscope) give different optical profile to both plagioclase feldspar and alkali feldspar. Our limited dataset was captured in a single stage rotation angle, hence the exclusion.

## III. OUR METHOD

This section details our proposed method of using edge and colour analysis to perform pixel-wise classification of rock thin section image as either quartz or accessory minerals.

### A. *Image Acquisition*

The study of rock minerals and textures is called petrography [12]. Thin section petrography is the study of microscopic features using a "polarizing" or "petrographic" microscope. A thin section is a 30 μm (= 0.03 mm) thick slice of rock attached to a glass slide with epoxy. Thin section reveals the textures and distributions of mineral. At this thickness level, most minerals become transparent thus allowing it to be studied under a microscope using transmitted light.

In this study, we used a total of 60 digital images taken from 20 thin sections. Thin sections are taken from four types of intrusive igneous rocks: Diorite, Tonalite, Granite, and Adamellite. All images are stored with a dimension of 2048 ×1536 pixels with 96 dpi resolution in RGB format. Each image is converted into two different formats: (i) grey scale 8-bit image with 512 x 384 pixels resolution, and (ii) RGB colour image with 512 x 384 pixels resolution.

The grey scale image is used with Canny edge detector [16] for quartz classification. Quartz are colorless and has none or minimal cleavages hence the reason for greyscale image and edge analysis. The colourful nature of accessory mineral enable us to use the RGB image to generate color histogram.

### B. *Feature Extraction and Analysis*

#### a) *Edge Image*

An edge filter is applied to the grey scale image to obtain an edge image. In this study, we used the Canny edge operator [16]

to perform the previously-mentioned task. According to [13], edge is a useful image feature since it represents the changes in local intensity. Since quartz has none or minimal amount of cleavages (compared to the other mineral types), we reason that an edge analysis would be useful.

*b) RGB Image*

We build colour histogram from the source 24 bits colour image of the rock thin section. To identify accessory minerals, two statistical measures, i.e., mean (see Equation 1) and variance (see Equation 2) are used,

$$\bar{x} = \frac{1}{n} \sum_{i=l}^{n} x_i \quad (1)$$

$$\hat{\sigma}_n^2 = \sum_{i=l}^{n} (x_i - \bar{x}) \quad (2)$$

A set of pixels is considered as belonging to the accessory mineral class if the variance of colors exhibited by the members of the set exceeds a threshold value. The variance is obtained from the built RGB histogram.

*C. Pixel-wise Classification*

Next, both grey scale and RGB image is partitioned into a finite number of cells that form a grid structure. We choose 4x4, 8x8, 16x16 and 32x32 grid cells. Figure 1 and 2, each shows a sample of a rock thin section image that is clustered into $m \times n$ grid cells. The cells are classified individually and the overall classification is achieved using a majority voting scheme, similar to [17], [18], [19], [20], [21] and [22].

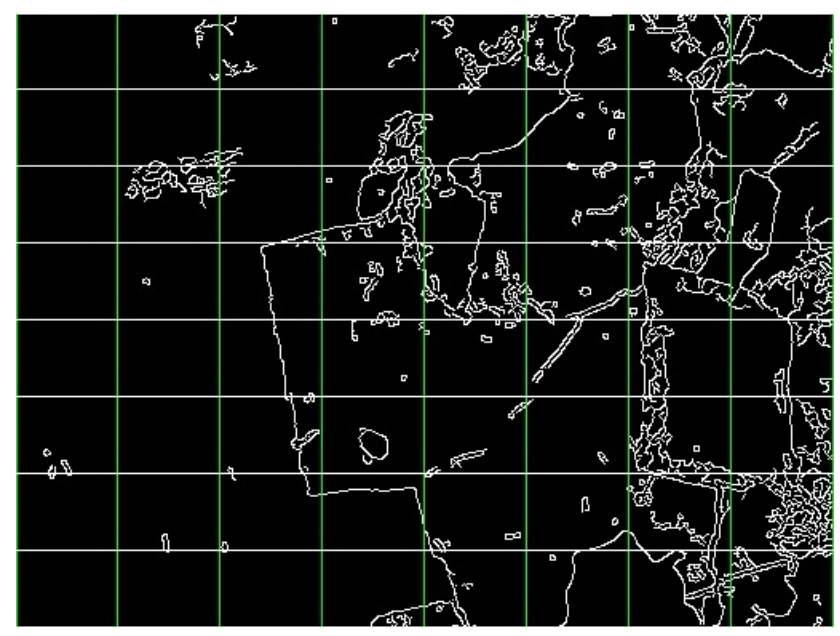

Fig. 1. An edge image (granite) divided into $8 \times 8$ grid cells.

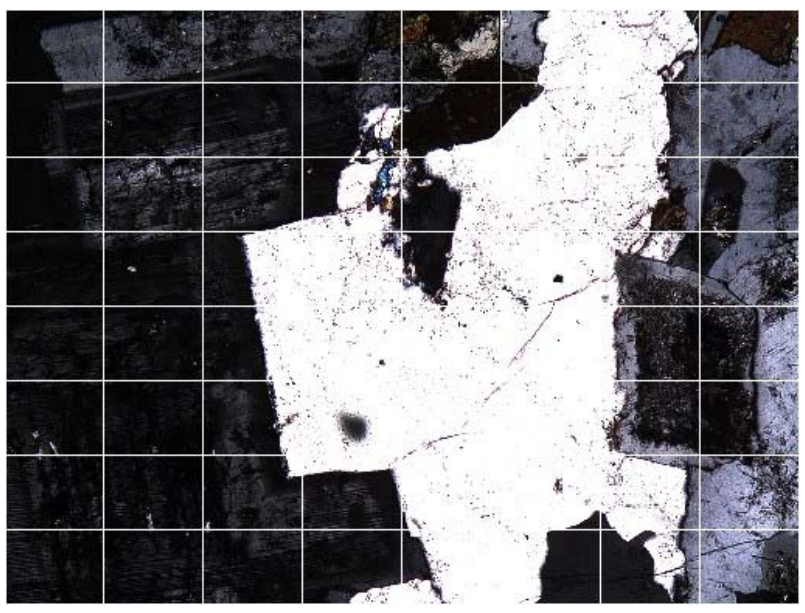

Fig. 2. An RGB image (granite) divided into $8 \times 8$ grid cells.

## IV. ANALYSIS

In the first experiment, images from each rock thin section samples are processed using multiple thresholds, generating a total of 960 images. Table 2 listed the thresholds used in the experiment. A sample positive output (cell grid size: 16x16, image resolution: 512x384, cell resolution: 32x24, cell count: 256) is shown in Table 4. Our method also produces the percentage of accessory minerals and quartz. Based on these percentages, see Table 1, the system automatically classifies the type of igneous rock for that particular sample.

TABLE 3. PARAMETER SET FOR THE FIRST EXPERIMENT

| Cell dimensions | Thresholds for Canny Operator | Thresholds for Colour Variance |
|---|---|---|
| 4 x 4 | 0.01, 0.02, 0.03 | 50, 100, 150, 200, 250, 300 |
| 8 x 8 | 0.01 | 50, 100, 150, 200, 250, 300 |
| 16 x 16 | 0.01 | 50, 100, 150, 200, 250, 300 |
| 32 x 32 | 0.01 | 50, 100, 150, 200, 250, 300 |

TABLE 4. SAMPLE OUTPUT

| **Sample of 8x8 analysis result** | |
|---|---|
| Opening granite/Diorite1.jpg | |
| Params | = 8x8 |
| Image resolution | 512 x 384 |
| Cell resolution | 64x48 |
| Number of cells | 64 |
| t_nonzero | 0.01 |
| Accessory Minerals | 17/64 (0.270000) |
| t_variance | 50 |
| Quartz | 0/64 (0.000000) |
| **It's a Diorite!** | |

TABLE 5. PARAMETER SET FOR THE SECOND EXPERIMENT

| Cell dimension | Threshold for Canny Operator | Threshold for Colour Variance |
|---|---|---|
| 4 x 4 | 0.01, 0.02, 0.03 | 100, 200, 300 |
| 8 x 8 | 0.01, 0.02, 0.03 | 100, 200, 300 |
| 16 x 16 | 0.01, 0.02, 0.03 | 100, 200, 300 |
| 32 x 32 | 0.01, 0.02, 0.03 | 100, 200, 300 |

In the second experiment, we chose 40 out of 60 images. The 40 images are selected randomly based on the correct percentage of classification. This time, the selected images are tested under a different threshold set, see Table 5. The new and finer threshold range is aimed to reveal the optimal threshold value within the best performing range of values from the first experiment.

From the classification results of the second experiment, we obtained the number of True Positives and False Positives. Using these values, we calculate the classification precision as,

$$\text{Precision} = \frac{\text{True Positives(TP)}}{\mathit{True\ Positives(TP)}\ + \mathit{False\ Positives\ (FP)}} \quad (3)$$

Based on the results, there are notable differences in classification performance between classes of rocks type. The rest of this section shall discuss these differences and offer analysis on the possible causes.

### *A. Class of Adamellite*

Adamellite is an intrusive, felsic, igneous rock, which typically is light coloured phaneritic (coarse-grained). The optimal parameter set for Adamellite is: 16x16 grid, Canny threshold between 0.01 to 0.02m and colour variance threshold between 100 to 200. Another optimal parameter set for Adamellite is: 32x32 grid, Canny threshold between 0.01 to 0.03, and colour variance threshold of 100. This is due to the non-homogenous grain size of mineral for Adamellite. The grain size is determined by the cooling mass of magma.

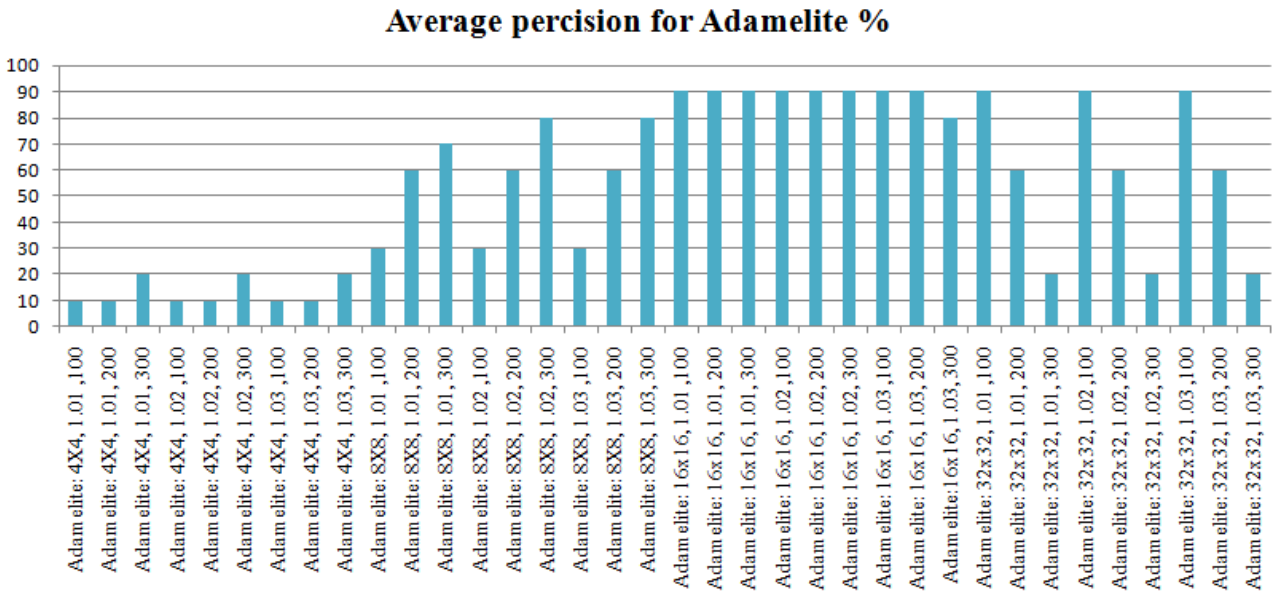

Fig. 3. Average precision for Adamellite.

### *B. Class of Granite*

Granite is hard, tough and widely used as a construction stone. It is a relatively light coloured, holocrystalline, usually coarse to medium-grained, phaneritic, plutonic igneous rock. Optimal parameter set is: 16x16 or 32x32 grid, a colour variance threshold of 300, with 90% and above precision value. The optimal Canny threshold ranges from 0.01 to 0.03 for grid size of 16x16, and from 0.01 to 0.02 for grid size of 32x32. We identify that the non-homogenous grain size of mineral in Granite is the main factor that effects the differences classification.

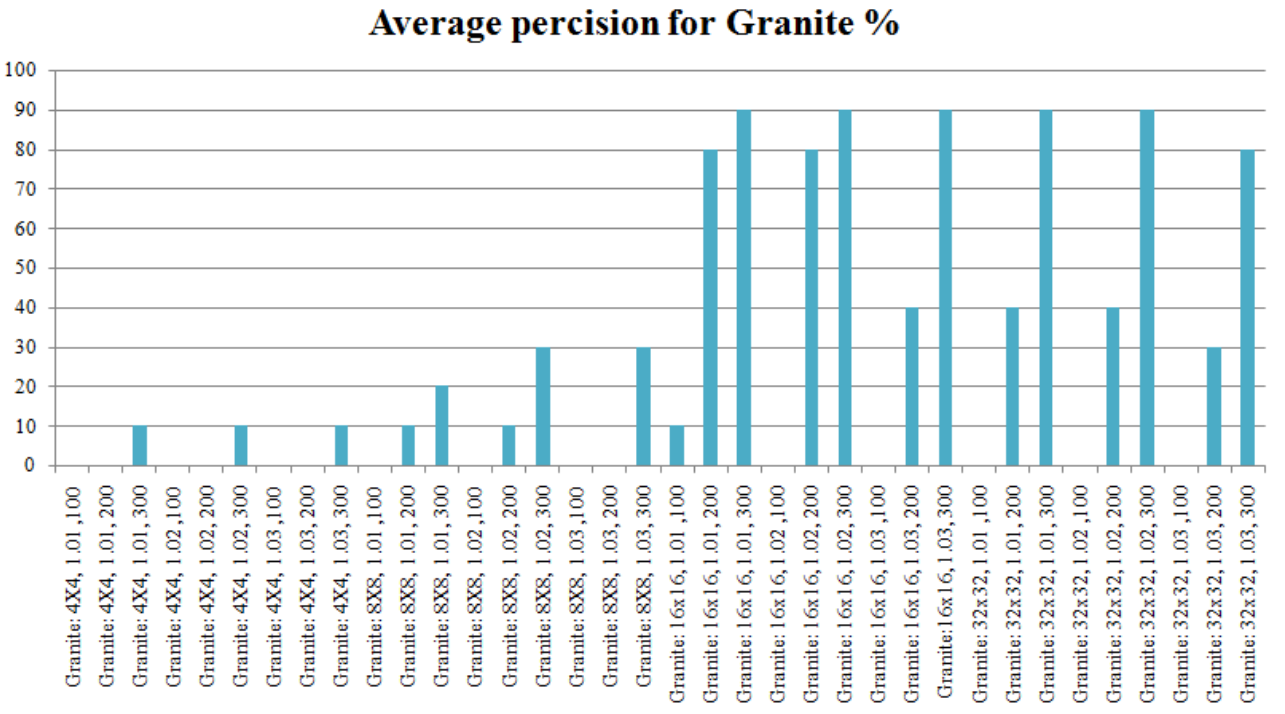

Fig. 4. Average precision for Granite.

### *C. Class of Diorite*

Diorite is a grey to dark-grey, coarse to fine-grained, intermediate intrusive igneous rock. Optimal parameter set for Class of Diorite consists of a wider range of grid size: 4x4, 8x8 and 16x16. For grid size of 16x16, the optimal threshold for Canny detector is in the range of 0.01 to 0.03, and the optimal colour variance threshold is 100. The findings for this class show the homogeneity of the grain size. Both quartz and accessory minerals possesses equal distribution in term of grain size.

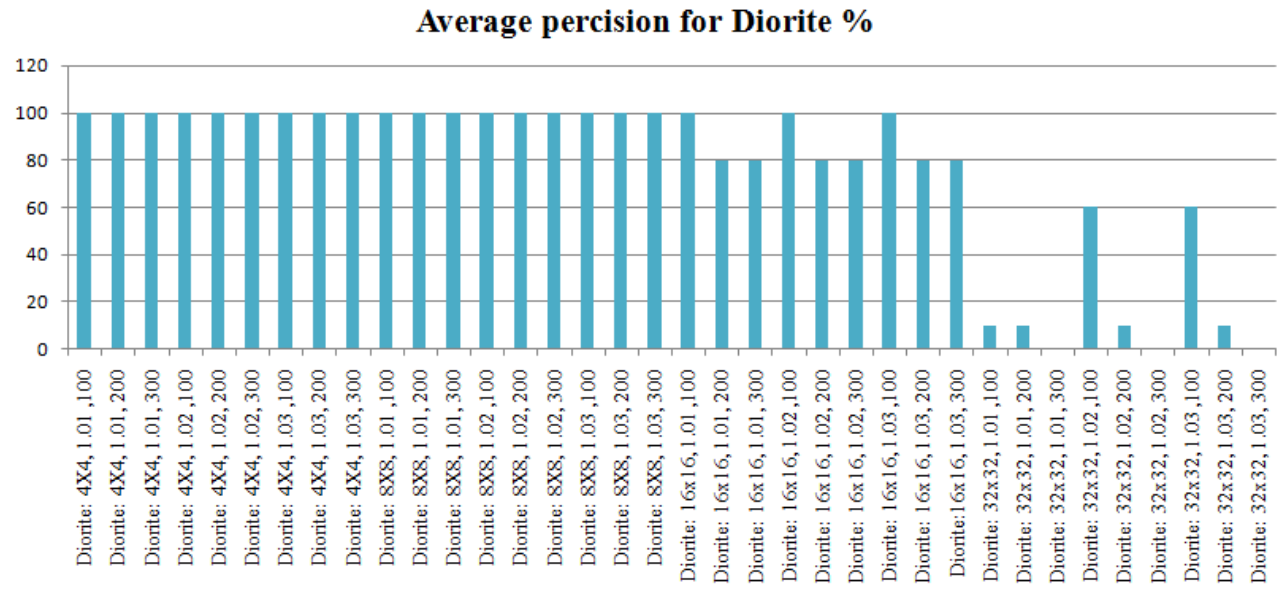

Fig. 5. Average precision for Diorite.

### *D. Class of Tonalite*

The optimal parameter set for Class Tonalite consists of 32x32 grid size, Canny threshold value between 0.01 to 0.03 and colour variance threshold of 100. Most of the combinations performed poorly for this mineral class. This is due to Tonalite only containing a small grain size of quartz and accessory minerals.

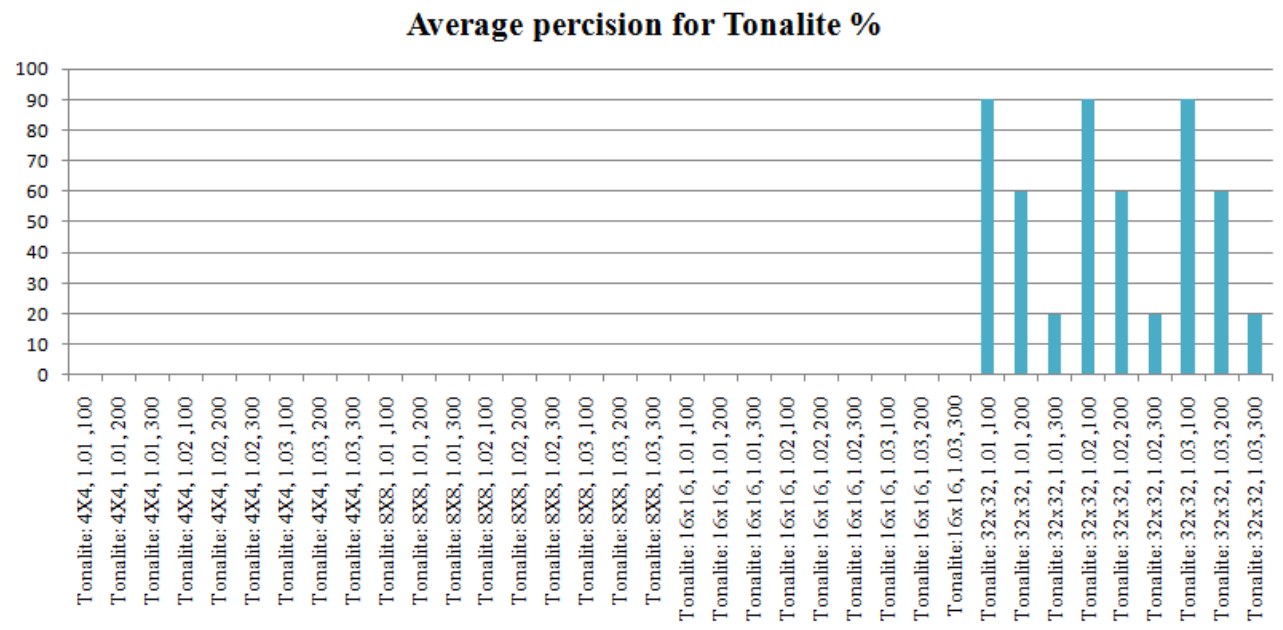

Fig. 6. Average precision of Tonalite.

We made several observations: (i) the smaller grid cell sizes can capture the subtle differences of pixels and performs best for minerals with small grain size of quartz and accessory minerals; (ii) classifications for minerals with medium grain size are more accurate using 16x16 and 32x32 grid cell size; (iii) classifications for minerals that has homogenous texture, are accurate by any grid cell sizes, however the best performing ones are 4x4, 8x8 and 16x16; (iv) Canny edge detector [16] is a capable method for finding edges in rock thin section images, allowing us to successfully detect the presence of monotone and cleavage-less quartz; and (v) the use of colour variance is sufficient to detect accessory minerals due to their colourful nature.

## V. CONCLUSION

In conclusion, the experimental results show that our method is capable of precise classification of igneous rocks based on the detection of quartz and accessory minerals. We exploit the visual properties of quartz and accessory minerals to design our pixel-wise classification method. A grid-based approach is employed to obtain local classifications which later are used in a majority voting scheme to determine the overall classification of the thin section image.

For future work, the followings are considered: (i) using a higher-dimensional image descriptor to capture the mineral visual characteristics; and (ii) include discriminative features learned from plagioclase and alkali feldspar to further improve the classification result.

## ACKNOWLEDGMENT

The authors wish to acknowledge the Malaysian Ministry of Higher Education (MOHE) and Universiti Malaysia Sarawak (UNIMAS). Also, the authors would like to express their gratitude for the kind permission given to them to use related services and facilities at the Lapidary Lab, Mineralogy and Petrology Lab, at the Department of Minerals and Geoscience Malaysia, Sarawak.